\documentclass[acmtog]{acmart}
\AtBeginDocument{%
  }

\setcopyright{cc}
\setcctype{by}
\acmJournal{TOG}
\acmYear{2025} \acmVolume{44} \acmNumber{6} \acmArticle{189}
\acmMonth{12} \acmPrice{}\acmDOI{10.1145/3763300}

\usepackage{pifont}

\usepackage{multirow} 
\usepackage{url} 
\usepackage{caption,subcaption,booktabs}
\usepackage{xcolor,colortbl}
\usepackage{cuted} 
\usepackage{capt-of}
\usepackage[font={small}]{caption}
\usepackage{epigraph}
\usepackage{comment}
\usepackage{enumitem}
\usepackage{algorithm}
\usepackage{algpseudocode}
\usepackage{todonotes}

\newcommand{\ra}[1]{\renewcommand{\arraystretch}{#1}} 

\begin{document}

\title{Generative Head-Mounted Camera Captures for Photorealistic Avatars}

\author{Shaojie Bai$^{1}$}
\authornote{Both authors contributed equally to this research.}
\email{shaojieb@gmail.com}
\author{Seunghyeon Seo$^{1,2}$}
\authornote{Work done during an internship at Meta Reality Labs.}
\authornotemark[1]
\email{zzzlssh@snu.ac.kr}
\author{Yida Wang$^{1}$}
\email{yidawang@meta.com}
\author{Chenghui Li$^{1}$}
\email{leochli@meta.com}
\author{Owen Wang$^{1}$}
\email{ojwang@meta.com}
\author{Te-Li Wang$^{1}$}
\email{teli@meta.com}
\author{Tianyang Ma$^{1}$}
\email{tianyang@meta.com}
\author{Jason Saragih$^{1}$}
\email{jsaragih@meta.com}
\author{Shih-En Wei$^{1}$}
\email{swei@meta.com}
\author{Nojun Kwak$^{2}$}
\email{nojunk@snu.ac.kr}
\author{Hyung Jun Kim$^{1}$}
\email{johnkim@meta.com}
\affiliation{%
  \institution{\\ $^{1}$Meta Reality Labs}
  \country{USA}
  \institution{$^{2}$Seoul National University}
  \country{Republic of Korea}
}
\settopmatter{printacmref=false}

\renewcommand{\shortauthors}{Bai et al.}

\begin{abstract}
  Enabling photorealistic avatar animations in virtual and augmented reality (VR/AR) has been challenging because of the difficulty of obtaining ground truth state of faces.
  It is \emph{physically impossible} to obtain synchronized images from head-mounted cameras (HMC) sensing input, which has partial observations in infrared (IR), and an array of outside-in dome cameras, which have full observations that match avatars' appearance.
  Prior works relying on analysis-by-synthesis methods could generate accurate ground truth, but suffer from imperfect disentanglement between expression and style in their personalized training. 
  The reliance of extensive paired captures (HMC and dome) for the \emph{same} subject makes it operationally expensive to collect large-scale datasets, which cannot be reused for different HMC viewpoints and lighting.
  In this work, we propose a novel generative approach, \textbf{Gen}erative \textbf{HMC} (GenHMC), that leverages \emph{large unpaired HMC captures}, which are much easier to collect, to directly generate high-quality \emph{synthetic} HMC images given any conditioning avatar state from dome captures. 
  We show that our method is able to properly disentangle the input conditioning signal that specifies facial expression and viewpoint, from facial appearance, leading to more accurate ground truth.
  Furthermore, our method can generalize to unseen identities, removing the reliance on the paired captures.
  We demonstrate these breakthroughs by both evaluating synthetic HMC images and universal face encoders trained from these new HMC-avatar correspondences, which achieve better data efficiency and state-of-the-art accuracy.
\end{abstract}

\ccsdesc[500]{Computing methodologies~Computer vision}
\ccsdesc[500]{Computing methodologies~Animation}
\ccsdesc[500]{Computing methodologies~Generative and developmental approaches}
\ccsdesc[500]{Computing methodologies~Neural networks}
\ccsdesc[500]{Computing methodologies~Learning paradigms}

\keywords{Diffusion models, synthetic data, digital human, real-time 3D facial animation}
\begin{teaserfigure}
  \includegraphics[width=\textwidth]{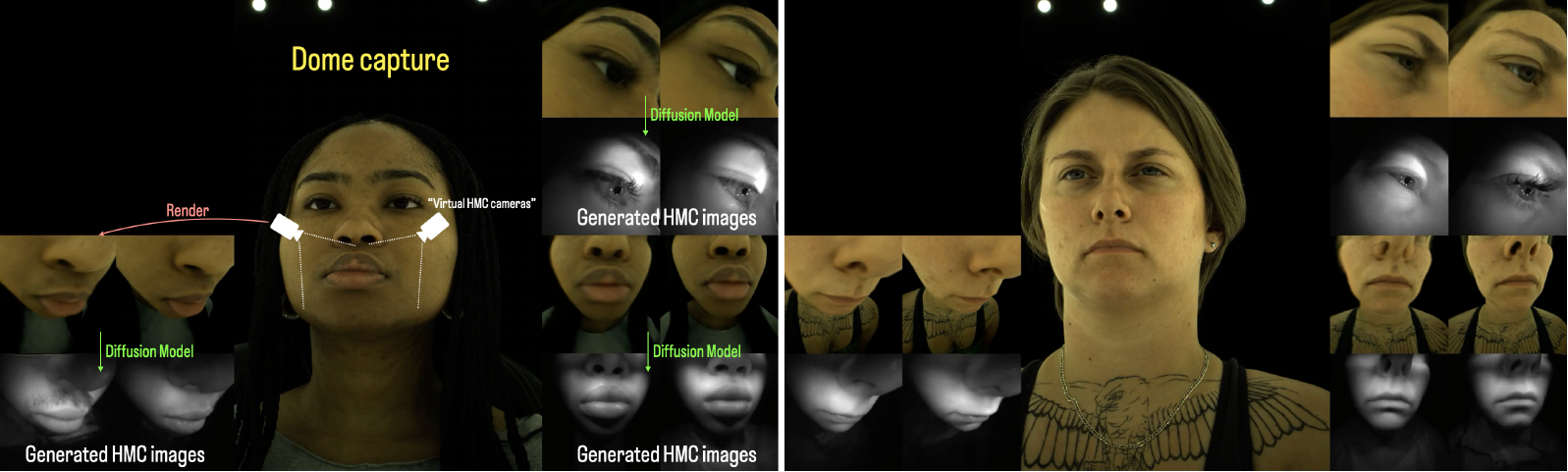}
  \caption{GenHMC establishes an HMC-avatar correspondence \emph{directly} via a dome capture and an image diffusion model. These ``generative'' captures introduce natural augmentations to identity information without altering the expressions.
  }
  \label{fig:teaser}
\end{teaserfigure}


\maketitle

\section{Introduction}

The emergence of Virtual Reality (VR) devices made it possible to create a system that faithfully simulates real-world human interactions. Such a system requires not just high-quality photorealistic avatars, but also the ability to animate them to emulate the \emph{immediacy} and \emph{fidelity} of face-to-face conversations. Recent works have made huge progress towards this goal of photorealistic telepresence~\citep{wei2019VR,bai2024ue}.

Despite the limited coverage of head-mounted cameras (HMC) sensing input, direct regression of avatar states (\textit{e.g.}, blendshapes or learned latent codes) from HMC images have been proven effective and efficient~\citep{bai2024ue}. The core challenge lies in the establishment of abundant input-output pairs of the system, known as the HMC-avatar correspondence problem.
The difficulty comes from the opaque nature of VR headsets: it is \emph{physically impossible} to simultaneously capture full face with HMCs, which generates sensing inputs, and with an array of outside-in cameras, where images could be used to generate ground truth state of the avatar. 
To overcome this challenge, prior works~\citep{wei2019VR,schwartz2020eyes} adopted analysis-by-synthesis approaches which require (1) capturing the same subject in both HMCs and in a multi-camera dome; (2) building avatars with the data including the subject; (3) training style-transfer models to bridge domain gaps in lighting direction and spectrum (\textit{i.e.}, infrared vs. visible), and (4) fitting full avatars to partial HMC observations.

This process has several key drawbacks in terms of both operational cost and quality. On the cost side, the need for paired HMC and dome captures creates inherent limitations. When a new headset is introduced, the dataset must be recreated from scratch rather than reusing existing dome captures. Additionally, the analysis-by-synthesis training must be redone whenever the avatar’s expression space definition is updated.
On the quality side, challenges arise from imperfect disentanglement between style transfer and expressions, leading to the "cheating" effect described in~\citet{wei2019VR}, due to the large domain gap between HMC and dome images. Fitting a complete avatar onto partial HMC observations also requires proper regularization for unseen parts (\textit{e.g.}, hair) to prevent out-of-distribution artifacts, which can reduce fitting accuracy. Furthermore, the precision of the process is constrained by the avatar’s expressivity range, which is assumed to encompass all facial expressions in the HMC images. Any discrepancies in appearance or shape (\textit{e.g.}, makeup) between the dome and HMC images can lead to fitting inaccuracies.

In this work, we introduce \textbf{Gen}erative \textbf{HMC} (GenHMC), a novel solution to the HMC-avatar correspondence problem. Instead of optimizing avatar's expression to match real HMC images, we present a method to leverage generative models to realize ``the other way around'': putting virtual HMCs on avatars to synthesize HMC images, with their expression captured in dome, via diffusion models. 
To achieve this, we leverage a large pool of \emph{unpaired} HMC captures (which are much easier to obtain than paired data) to build a foundational HMC image generation model.
Specifically, we design and train a U-Net based diffusion model, conditioned on expression-related features, such as keypoints and segmentations, which are obtainable from both HMC images and avatar projections.
With this model, we can now freely sample new HMC images from any specific avatar state captured in a dome.
Kindly note that our method focuses on establishing correspondences between expressions and HMC images, rather than achieving full appearance-level correspondence.
In that sense, the ``avatar-correspondence problem'' is addressed specifically from the perspective of training expression encoders, where our goal is to preserve expression fidelity while allowing for appearance diversity.

Our approach effectively addresses several of these limitations.
First, it significantly reduces the cost of generating HMC-avatar correspondences: our diffusion model requires training only once, and generalizes to expression features from avatars with unseen identities. This eliminates the need for paired captures, and repeated training for any updates in the avatar's expression space, as the expression features are agnostic to latent expression values.
Second, the quality of the correspondences is improved by directly using ground truth measurements from dome captures, free from out-of-distribution issues.
When sampling with a certain expression captured in the dome, the model can generate diversity in the non-expression factors (\textit{e.g.}, illumination, appearance), improving the robustness of downstream face encoder systems.
Finally, with extensive experiments, we show that combining these synthetic data with conventional HMC-correspondences (from real HMC images) significantly outperforms the baseline face encoder trained solely on real data.
Our contributions are summarized as follows:
\begin{itemize}[noitemsep,topsep=0pt,parsep=0pt,partopsep=0pt, leftmargin=*]
\item A novel approach to generate HMC-avatar correspondences by sampling from an expression-conditioned diffusion model trained from unpaired HMC captures.
\item A demonstration that face encoders trained from synthetic data outperform baseline models.
\item A scaling law analysis showing how GenHMC impacts the performance and efficiency of the face encoder training pipeline.
\end{itemize}

\section{Related Works}
\subsection{Photorealistic Avatars}
The pursuit of photorealistic digital avatars has become central in computer vision and graphics, driven by the demand for immersive virtual experiences. Foundational work by~\citet{blanz1999morphable} introduced 3D morphable models for realistic face synthesis, followed by multi-view capture systems that improved facial geometry and appearance reconstruction~\citep{zhang2004spacetime,klaudiny2012high,huang2006subspace,fyffe2014driving,furukawa2009dense,ghosh2011multiview,bickel2007multi,beeler2011high,bradley2010high}. Texture mapping further enhanced realism with lifelike skin detail~\citep{debevec2000acquiring,weyrich2006analysis,donner2008layered}.
More recently, the field has shifted to learning-based methods. Building on Active Appearance Models~\citep{cootes2001active}, deep learning approaches~\citep{Gecer_2019_CVPR,lombardi2018deep,Shysheya_2019_CVPR} have achieved higher fidelity and expressiveness than classical techniques~\citep{ichim2015dynamic,cao2016real,seymour2017meet}. These methods support advanced capabilities like gaze control, relighting, and volumetric modeling for hair~\citep{schwartz2020eyes,bi2021deep,yang2023towards,rosu2022neural}, and enable realistic avatar generation from limited inputs, including handheld device captures~\citep{cao2022authentic,grassal2022neural}.

\subsection{Establishing HMC-avatar Correspondences}
Registering 3D face models to HMC captures is essential for accurate VR face tracking. Traditional landmark-based methods~\citep{wei2019VR} yield sparse correspondences and struggle with subtle motions under oblique views. Depth sensor-based methods~\citep{Hao2015,FaceVR2018} captured lower-face geometry more directly but required carefully positioned cameras. Some approaches relied on users mimicking predefined expressions aligned with audio, resulting in coarse, discrete tracking~\citep{Olszewski2016}.
With more expressive avatars, analysis-by-synthesis methods~\citep{Nair2008} were explored, but domain gaps between avatar renderings and real HMC images remained a challenge. \citet{wei2019VR} addressed this with a two-stage CycleGAN and differentiable rendering pipeline, though semantic shifts often occurred. \citet{schwartz2020eyes} improved on this by jointly training expression regression and style transfer, reducing domain mismatch through end-to-end learning.
In this work, we leverage high-fidelity dome captures and a generative diffusion model to establish accurate, expressive HMC-avatar correspondences with far less manual effort and system complexity.

\subsection{Synthetic Data Training for Neural Network}
Synthetic data has become a pivotal strategy in generating large labeled datasets, enhancing vision task performance~\citep{tian2024stablerep,hesynthetic}.
It has been widely utilized in various computer vision domains~\citep{peng2015learning,ros2016synthia,chen2019learning,abu2018augmented,fischer2015flownet,mayer2016large}, effectively addressing data scarcity and annotation costs.
Recently, large-scale text-to-image models~\citep{rombach2022high,nichol2022glide,ramesh2022hierarchical,balaji2022ediff,saharia2022photorealistic} have expanded synthetic data capabilities. Integrating such generated images with real datasets substantially improves supervised learning tasks~\citep{azizi2023synthetic,hesynthetic,sariyildiz2023fake}.
Notably, synthetic images from GLIDE~\citep{nichol2022glide} enhanced zero-shot and few-shot classification~\citep{hesynthetic}, and diffusion-based augmentation improved few-shot learning~\citep{trabucco2023effective}.

\section{Preliminaries}
\subsection{Codec Avatars}
A Codec Avatar, which consists of an encoder-decoder pair, denoted as $(\mathcal{E}, \mathcal{D})$, respectively, has emerged as a powerful paradigm for compactly representing and reconstructing high-fidelity facial dynamics.
It aims to reduce both reconstruction error (distortion) between the original and reconstructed data, and the overall delay (latency) involved in replicating a user’s appearance and behavior at a remote location.
In this paper, we use a universal facial encoder that takes image data from VR headset cameras as input and generates encodings of the user's state in a universal latent space, which is independent of their identity~\citep{bai2024ue}.

The universal facial encoders are based upon the universal prior models (UPMs) for faces~\citep{cao2022authentic}.
Its Codec Avatar framework includes an expression encoder, $\mathcal{E}_\text{exp}$ and separate hypernetwork identity encoder, $\mathcal{E}_\text{id}$.
$\mathcal{E}_\text{exp}$ generates expression latent codes $\mathbf{z} \in \mathbb{R}^{4 \times 4 \times 16}$.
This convolutional latent space ensures that each latent dimension influences specific spatial regions while maintaining semantic consistency across different identities.
$\mathcal{E}_\text{id}$ processes a neutral texture map $\mathbf{T}_\text{neu}$ and a neutral geometry image $\mathbf{G}_\text{neu}$ to generate multi-resolution bias maps $\Theta_\text{id}$ serving as parameters for the decoder $\mathcal{D}$, which reconstructs volumetric slabs $\mathbf{M}$ and geometry $\mathbf{G}$ based on disentangled control inputs, including viewpoint $\mathbf{v} \in \mathbb{R}^{6}$, gaze directions $\mathbf{g} \in \mathbb{R}^{6}$, and expression latent code $\mathbf{z}$.
The final image $I_\text{vol}$ is rendered by ray-marching $R_\text{vol}$.
Furthermore, additional network modules are added to $\mathcal{D}$ to estimate the texture maps $\mathbf{T}$ for the face and eyes, allowing the final mesh $I_\text{mesh}$ to be obtained using a mesh renderer $R_\text{mesh}$.
Kindly refer to \citet{bai2024ue} for further details.

\subsection{Generative Diffusion Model}
Diffusion models~\citep{sohl2015deep,ho2020denoising,song2020score,song2019generative} have been introduced as powerful generative frameworks and are now widely used for high-quality image synthesis in computer vision. They transform data into noise via a forward process and learn to reverse it through a parameterized backward diffusion. The generation process can also be conditioned on various inputs such as text, keypoints, or class labels~\citep{rombach2022high,choi2021ilvr,saharia2022photorealistic,zhang2023adding,kawar2023imagic,hertz2022prompt,hu2024animate}.

A diffusion model is trained by minimizing a variational upper bound or a simplified loss.
A common practice is to predict the noise $\varepsilon \sim \mathcal{N}(\mathbf{0}, \mathbf{I})$ added at step $t \sim \mathcal{U}([0,1])$ and the corresponding training objective is as follows:
\begin{equation}
  \mathbb{E}_{\mathbf{x}, \mathbf{\varepsilon}, \mathbf{c}, t} \left[ ||\varepsilon_{\theta}(\alpha_{t}\mathbf{x} + \sigma_{t}\varepsilon, \mathbf{c}, t) - \mathbf{\varepsilon}||^{2}_{2} \right],
  \label{eq:diffusion-objective}
\end{equation}
where $\varepsilon_{\theta}$ and $\mathbf{c}$ are the diffusion model (with $\varepsilon$-prediction) and the condition, respectively.
Typically, $\alpha_{t}$ is chosen such that it gradually decreases with $t$, preserving less of the original signal over time, while $\sigma_{t}$ increases with $t$, adding more noise.
Therefore, the term $\alpha_{t}\mathbf{x} + \sigma_{t}\varepsilon$ represents the noisy version of $\mathbf{x}$ at time $t$ with $\alpha_{t} = \sqrt{\bar{\alpha}_{t}}$ and $\sigma_{t} = \sqrt{1 - \bar{\alpha}_{t}}$ derived from a variance schedule, where $\bar{\alpha}_{t} = \prod_{s=1}^{t}(1-\beta_{s})$ is the total noise variance.
Note that $\beta$ denotes the small noise variance added at each step, and kindly refer to \citet{ho2020denoising} for more details.

Furthermore, a number of sampling strategies have been actively explored recently~\citep{ho2020denoising,songdenoising,phillips2024particle,liu2022pseudo,lu2022dpm,watson2022learning}.
Among them, DDPM~\citep{ho2020denoising} and DDIM~\citep{songdenoising} both begin by drawing a sample from pure noise, $\varepsilon_{T} \sim \mathcal{N}(\mathbf{0}, \mathbf{I})$, and iteratively produce a sequence of points,
$\varepsilon_{T},\varepsilon_{T-1}, \cdots, \varepsilon_{1}$, whose level of noise is gradually reduced throughout $T$ iterations.

Classifier guidance~\citep{dhariwal2021diffusion} uses gradients from a pre-trained model to boost sample quality in conditional diffusion models with reducing diversity.
Classifier-free guidance~\citep{ho2022classifier}, on the other hand, avoids relying on a separate pre-trained model.
Instead, it trains a single diffusion model for both conditional and unconditional objectives by randomly dropping the condition $c$ during the training phase.
When sampling, it adjusts the predicted noise as follows:
\begin{equation}
  \tilde{\varepsilon}_{\theta}(\varepsilon_{t}, \mathbf{c}, t) = w\varepsilon_{\theta}(\varepsilon_{t}, \mathbf{c}, t) + (1-w)\varepsilon_{\theta}(\varepsilon_{t}, t)
  \label{eq:cfg-sampling}
\end{equation}
where $\varepsilon_{\theta}(\varepsilon_{t},\mathbf{c}, t)$, $\varepsilon_{\theta}(\varepsilon_{t}, t)$, and $w$ denotes the conditional and unconditional $\varepsilon$-predictions, and guidance weight, respectively.
Setting $w>1$ strengthens its influence, whereas $w=1$ effectively disables classifier-free guidance.

\section{Method}
\label{sec:genhmc}
In this work, we propose GenHMC, a conditional diffusion framework that generates synthetic HMC captures using a combination of keypoint and segmentation, which we have named \text{keyseg map}, as the input condition.
This approach facilitates the straightforward establishment of correspondences between HMC and avatars using dome capture images (Sec.~\ref{subsec:hmc_diffusion}).
Moreover, it enables the generation of synthetic images with diverse natural augmentations for a single expression, enhancing both flexibility and usability.
Finally, we establish a system using the generative HMC captures created by GenHMC for universal facial encoder training, significantly improving the efficiency of data collection (Sec.~\ref{subsec:encoder_training}).
We also explore various applications of the proposed method, \textit{e.g.}, adapting to multi-view diffusion framework, light control, and glasses generation.
Kindly refer to our supplementary material for the related sections.

\begin{figure*}[thb!]
  \includegraphics[width=0.95\textwidth]{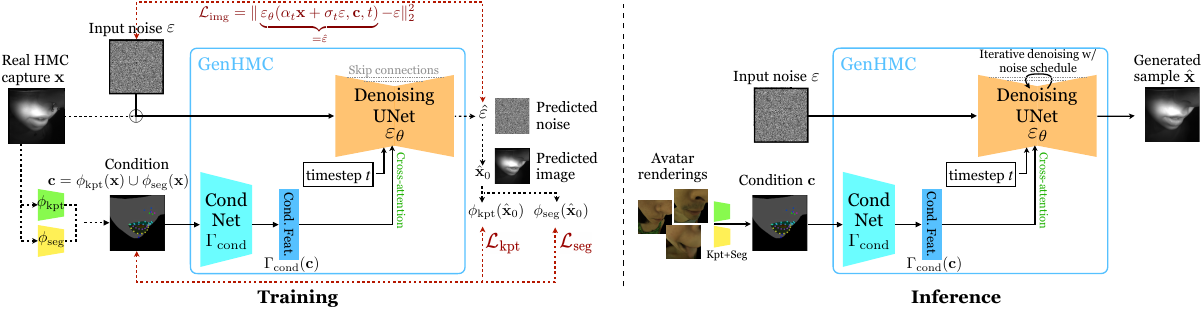}
  \caption{\textbf{The training and inference processes of our GenHMC model.} \textit{(Left)} During the training phase, we use the detected keypoint+segmentation annotations as the conditional signals, and self-supervise a conditional diffusion model training with the noise prediction loss as well as the perceptual losses. \textit{(Right)} At inference time, we obtain the keypoint+segmentation conditional as the (base) control signal for expressions, and sample the diffusion model to generate high-quality synthetic HMC data that aligns with this expression.
  }
  \label{fig:genhmc_overview}
\end{figure*}

\subsection{HMC Diffusion}
\label{subsec:hmc_diffusion}
An overview of the general GenHMC architecture is presented in Fig.~\ref{fig:genhmc_overview}. During training, we load a real, monochrome HMC image $\mathbf{x} \in \mathbb{R}^{256 \times 256}$, and apply it through a pre-trained face keypoint detection model $\phi_\text{kpt}$ and a face segmentation model $\phi_\text{seg}$. Specifically, the keypoint model detects for the upper and lower face, respectively:
\begin{itemize}[noitemsep,topsep=0pt,parsep=0pt,partopsep=0pt, leftmargin=*]
    \item eyebrows, outer/inner eyelids, and pupil center;
    \item upper/lower vermillion border, cupid's bow, upper/lower lip, left/right lateral commisure, left/right nostril, nasal base, nose tip, tongue tip, and the top/bottom of the chin,
\end{itemize}
each part with a different keypoint color. Likewise, the segmentation model detects for the upper and lower face, respectively:
\begin{itemize}[noitemsep,topsep=0pt,parsep=0pt,partopsep=0pt, leftmargin=*]
    \item face region, eyeball, iris and pupil;
    \item face region, upper/lower lips, inner mouth, and tongue,
\end{itemize}
each part with a different segmentation class color.
We then overlay the detected results of the two models on the same image $\mathbf{c}$:
\begin{align}
    \mathbf{c} = \mathbf{c}_\text{keyseg} = \phi_\text{kpt}(\mathbf{x}) \cup \phi_\text{seg}(\mathbf{x})
\end{align}
which we then use as the conditional signal for the generative model.
We chose to use an image combining segmentation maps and landmarks as the conditional input because this format enables precise alignment in the generation process, particularly around fine structures like pupils, lips, and facial contours.
Additionally, this image-based format allows for more intuitive editing and is flexible for future extensions.
As shown in our supplementary material, the same encoder architecture can be reused for other image-formatted conditions (\textit{e.g.}, abstract light maps), supporting broader applications.
Meanwhile, we randomly sample a Gaussian noise $\varepsilon \sim \mathcal{N}(0,I) \in \mathbb{R}^{256\times256}$, and timestep $t \in \mathcal{U}([0,1])$, and diffuse the original image $\mathbf{x}$ by $\tilde{\mathbf{x}}_t = \alpha_t \mathbf{x} + \sigma_t \varepsilon$ (following the notation of Eq.(\ref{eq:diffusion-objective}); see~\citet{ho2020denoising} for more details).

To train the diffusion model, we pass the conditional $\mathbf{c}$ through a shallow convolutional network $\Gamma_\text{cond}$, which is based on MobileNetV3~\citep{howard2019searching}, to obtain a lower-resolution feature map $\Gamma_\text{cond}(\mathbf{c}) \in \mathbb{R}^{C \times 12 \times 12}$ where $C$ is the feature dimension (in practice, we set $C=160$, following the output dimension of MobileNetV3). This conditional feature map is then flattened and injected into the bottleneck of the denoising U-Net (see \textbf{orange} module in Fig.~\ref{fig:genhmc_overview}) via a cross-attention layer. With the conditional feature map, the diffused image $\tilde{\mathbf{x}}_t$ and the timestep $t$, the denoising U-Net then tries to predict the noise $\hat{\varepsilon} = \varepsilon_\theta(\tilde{\mathbf{x}}_t, \mathbf{c}, t)$ and an posterior estimate of the reconstructed image $\hat{\mathbf{x}}_0$.

We propose three training objectives.
First, following prior works on diffusion models~\citep{song2020score,ho2020denoising}, the reconstruction loss on noise:
\begin{equation}
    \mathcal{L}_\text{img} = \|\varepsilon_\theta(\alpha_t \mathbf{x} + \sigma_t \varepsilon, \mathbf{c}, t) - \varepsilon\|_2^2
    \label{eq:genhmc-recon-loss}
\end{equation}
We also found it beneficial to add the perceptual losses which compares the originally detected keypoint heatmap activations, $\textsf{hm}(\phi_\text{kpt}(\mathbf{x}))$, with that of the reconstructed posterior, $\textsf{hm}(\phi_\text{kpt}(\hat{\mathbf{x}}_0))$,
\begin{equation}
    \mathcal{L}_\text{kpt} = \|\textsf{hm}(\phi_\text{kpt}(\mathbf{x})) - \textsf{hm}(\phi_\text{kpt}(\hat{\mathbf{x}}_0))\|_2^2
    \label{eq:genhmc-kpt-loss}
\end{equation}
and corresponding segmentation maps with a pixel-wise cross-entropy loss,
\begin{equation}
    \mathcal{L}_\text{seg} = \text{CrossEntropy}(\textsf{logit}(\phi_\text{seg}(\hat{\mathbf{x}}_0)), \phi_\text{seg}(\mathbf{x})),
    \label{eq:genhmc-seg-loss}
\end{equation}
where $\textsf{logit}(\cdot)$ represents the pre-softmax logits for the segmentation map. In summary, the total loss is a weighted combination of these three losses with the balancing weights $\lambda$s:
\begin{equation}
    \mathcal{L}_\text{total} = \lambda_\text{img} \mathcal{L}_\text{img} + \lambda_\text{kpt} \mathcal{L}_\text{kpt} + \lambda_\text{seg} \mathcal{L}_\text{seg}.
    \label{eq:genhmc-total-loss}
\end{equation}

At inference time, no real HMC image is available (after all, we are to generate them). We instead take the avatar renderings from simulated HMC viewpoints, detect/project the keypoint and segmentation maps on them, and use the outcome as conditional $\mathbf{c}$. In the meantime, we sample $\varepsilon \sim \mathcal{N}(0,I)$ and iteratively denoise it with $\varepsilon_\theta$ following a sampling schedule (DDPM~\citep{ho2020denoising}, DDIM~\citep{songdenoising}, etc.) from $t=1$ to $t=0$ with the conditional $\mathbf{c}$.

\textit{Discussions.} \,
We note that this flow falls into a broader category of works which employ conditional generative models as unsupervised priors to resolve \emph{inverse problems}~\citep{song2021solving,daras2024survey}, where the generative model aims to reconstruct a full observation from a partial measurement (in our case, keypoint+segmentation maps $\mathbf{c}$ representing the expression). We also specifically did not use the latent diffusion model (LDM)~\citep{rombach2022high} as it entails a separate VAE training process, and in this case we only need a relatively lightweight 256$\times$256 monochrome image.

\begin{figure}[thb!]
  \includegraphics[width=0.46\textwidth]{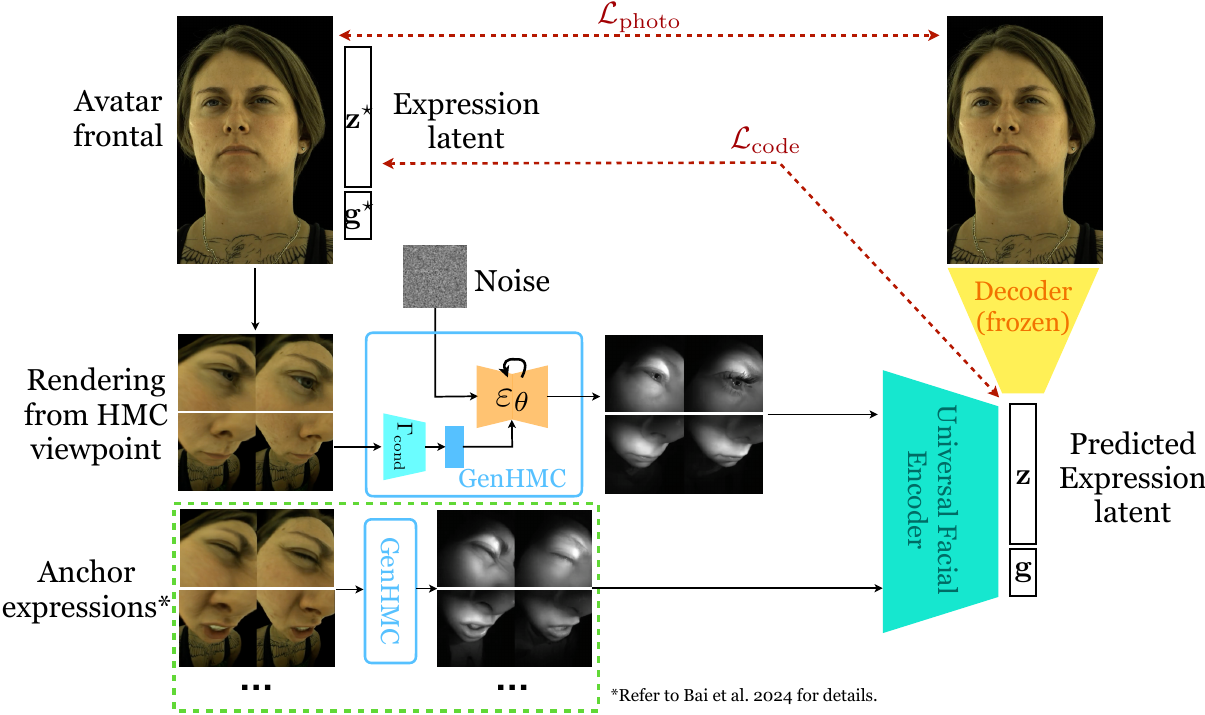}
  \caption{\textbf{Directly training universal facial encoder with GenHMC-based HMC-avatar correspondences}. Using the GenHMC approach, we can directly establish HMC-avatar correspondences and train on \emph{dome capture frames}. This is vastly different from the traditional approach which requires an HMC capture paired with the dome capture of the subject.
  }
  \label{fig:ue_training}
\end{figure}

\subsection{Encoder Training}
\label{subsec:encoder_training}
As illustrated in Fig.~\ref{fig:ue_training}, we present a novel encoder training system that minimizes the need for real HMC captures. The training system comprises two major components: GenHMC inference as illustrated in Fig.~\ref{fig:genhmc_overview}, and the encoder training with therefore synchronized dome assets.
For the first time, dome assets, such as ground truth expression codes, or even dome images can be directly utilized to train face encoders for VR headsets. This approach offers several key benefits:
(1) more accurate ground truth supervision from multi-view dome captures, comparing against pseudo correspondences established by \citet{wei2019VR}, (2) reduced need for paired HMC and dome captures, and (3) increased diversity in HMC inputs resulted from GenHMC.

We follow the exact setup of~\citet{bai2024ue} for the state-of-the-art universal facial encoder model architecture, and refer interested readers to it for more details.

\section{Experiments}

In this section, we demonstrate the quantitative and qualitative results of the GenHMC method. In Sec.~\ref{subsec:experiment-datasets}, we start by describing the datasets we use as well as the baseline approaches for establishing and training with the HMC-avatar correspondences.
Then in Sec.~\ref{subsec:experiment-analysis}, we delve into analysis of the scalability of GenHMC generative model, and study a key downstream application of the GenHMC framework by building a universal facial encoder system with the synthetic captures and comparing with the baseline method.

\subsection{Experimental Details}
\label{subsec:experiment-datasets}
\subsubsection{Datasets}
The datasets we use in this section consist of four different categories.

\textit{Avatar dome captures.} \,
We captured 341 subjects' multi-view face performances in a spherical capture dome equipped with 90 colored/monochrome high-resolution cameras. Specifically, we follow the setup of~\citet{cao2022authentic}, where each subject is asked to perform 20 minutes of facial expressions, neutral and emotional speeches, head movements, and gaze sequences. The visual data are captured with relighting at 30 FPS and resolution 4096$\times$2668, which is more than 12M frames in total. Subsequently, we use them to train a relightable universal Gaussian Codec Avatar model~\citep{li2024uravatar} (\textit{i.e.}, the \emph{decoder}) for the 341 subjects.

\textit{\textcolor{olive}{Paired} (with dome capture) HMC captures.} \,
For each of these 341 subjects, we also collect the head-mounted camera (HMC) captures of these very same subjects, on the same day, with a modified version of the Quest Pro VR headset that contains augmented training cameras which provide better lower-face visibility (we refer interested readers to Fig. 2 of~\citet{bai2024ue} for more details). Each subject is asked to perform roughly 9 minutes of range of facial motions, speeches, as well as gaze sequences, while changing the headset donning after every few capture segments. The HMC images are captured at 72 FPS with resolution 400$\times$400, but we downsample the images to 192$\times$192 for universal facial encoder training. This dataset consists entirely of \emph{real} HMC data, and we follow~\citet{wei2019VR}'s personalized style-transfer approach, which requires the augmented camera images, to generate the corresponding expression latent codes using the decoder.

\textit{\textcolor{orange}{Unpaired} large-scale HMC captures.} \,
We also leverage a large pool of HMC captures of over 18K subjects. The capture script is the same as the HMC captures above (roughly 40K frames per subject), except that none of these subjects has a corresponding dome capture (\textit{i.e.}, we don't have their avatars). This dataset is primarily used to train our GenHMC model. Note that while we are unable to release its detailed demographic statistics due to proprietary constraints, we would like to note that the dataset was collected with covering a diverse population across various attributes.

\textit{Synthetic HMC captures.} \,
For each dome capture frame of every subject, we render the photorealistic avatar from an HMC-alike camera view (with simulated distortions) and apply the GenHMC approach to generate the corresponding monochrome, infrared-spectrum synthetic HMC images following the process of Fig.~\ref{fig:genhmc_overview}. Therefore, these synthetic HMC captures have the same frame rate, frame count, and ``capture script'' as the aforementioned dome captures, totaling over 12M frames.

\subsubsection{Baselines}
Our method introduces a novel formulation for generating HMC-avatar correspondences. As the first generative approach that enables this, GenHMC allows many new functionalities and capabilities that had not existed in prior works, which we elaborate on in our supplementary material.

Meanwhile, we demonstrate the benefits of our approach by comparing the universal facial encoder trained using our GenHMC data, with the encoder trained using only real HMC data and the \emph{best} existing approach that builds pseudo HMC-avatar correspondences, requiring operationally costly paired dome \& HMC captures, augmented headset cameras, and person-specific style-transfer modules~\citep{wei2019VR}.
Unless otherwise noted, all encoders are initialized with pre-trained weights from the self-supervised learning (SSL) framework proposed by~\citet{bai2024ue}, which currently represents the state-of-the-art in encoder training, achieving the best generalization error on unseen identities.

\begin{figure*}[tbp!]
  \centering
  \includegraphics[width=0.95\textwidth]{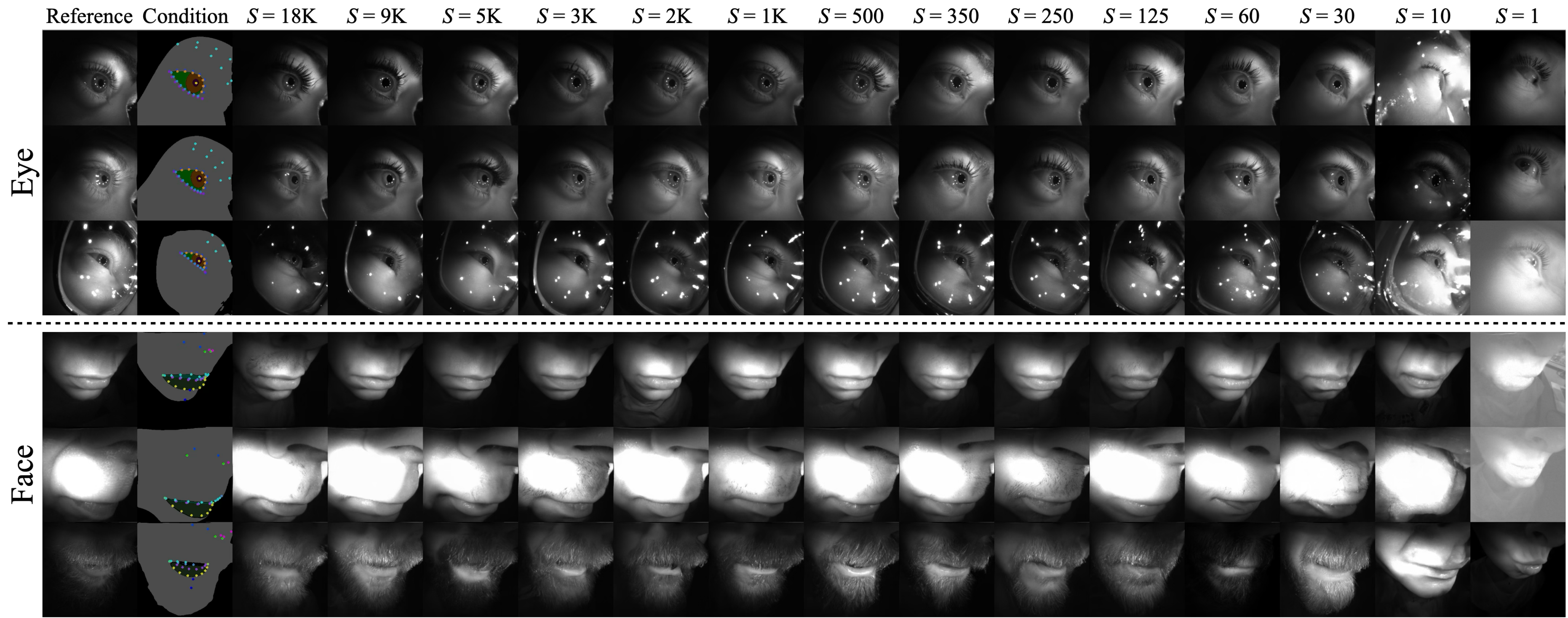}
  \caption{\textbf{Effect of training size on GenHMC inference results.}
  When the number of training subjects is small, the diversity of the generated output decreases, which can lead to degenerate results when an arbitrary keyseg map is provided as a condition.
  $S$ denotes the training dataset size, \textit{i.e.}, the number of training subjects.}
  \label{fig:eye_face_inference}
\end{figure*}

\subsubsection{Evaluation Metrics}
\label{subsubsec:evaluation-metrics}

To properly evaluate the universal facial encoder's generalizations, we keep a held-out set of 34 test subjects (out of 341) whose neither 3D avatar models nor HMC images are available during encoder training. We train on the other 307 subjects' real HMC data and/or GenHMC data (generated from their avatars), and evaluate the trained encoders on this held-out test set to assess the model's ability to generalize to unseen subjects. Kindly note that the keypoint/segmentation maps are used only as conditioning inputs for the diffusion model, not for determining correspondences, and the evaluation is performed against the dome-derived expression codes.

Following~\citet{bai2024ue}, we primarily compute the following set of metrics:
\begin{itemize}[noitemsep,topsep=0pt,parsep=0pt,partopsep=0pt, leftmargin=*] 
    \item \emph{Photometric Image $L_1$ Error}. Pixel-level errors of the renderings of the encoder-predicted expression on the avatar.
    \item \emph{Geometric $L_2$ error}. $L_2$ displacement error of the encoder-predicted vs. ground truth face mesh, measured in millimeters (mm) and averaged over all facial region vertices.
    \item \emph{Geometric $L_2$ error on mouth region}. Similar as the geometric $L_2$ error above, but only on the mouth (and surrounding) region vertices.
    \item \emph{Lip shape error}. The average of the horizontal (distance between left and right lip corners) and vertical (distance between center upper and lower lips) lip shape errors, which is especially important for speech.
\end{itemize}
Importantly, we note that throughout this section, the so-called ``ground truth'' expressions of the real HMC captures are not the true actual ground truth (which is impossible to obtain), but instead from the style-transfer based HMC-avatar correspondences generated by the method of~\citet{wei2019VR}. \textbf{It is therefore a biased estimation that is subject to errors still, but nevertheless a consistent reference in an ablative setting}, aligning with the prior works' convention~\citep{chen2021high,schwartz2020eyes,bai2024ue,martinezcodec}.

\subsection{Scalability of GenHMC}
\label{subsec:experiment-analysis}
To assess the impact of training data size on GenHMC's performance, we perform ablation studies by training models using subsets of the full dataset (18K subjects) with varying numbers of subjects (9K, 5K, 3K, and so on).  As expected, we observe that reducing the number of training subjects leads to a decrease in the quality and diversity of the generated images, as shown in Fig.~\ref{fig:eye_face_inference}.
Specifically, the inference results become increasingly degenerate, and the pixel-wise alignment between the generated outcome and the keypoint/segmentation maps degrades. In the extreme case where we train with only one subject, the model tends to generate images that closely resemble the attributes of that specific subject, indicating a significant loss of diversity.  These ablation studies demonstrate the importance of a sufficiently large and diverse unpaired dataset for GenHMC to generate high-quality and diverse images, although we note that comparable results can be achieved with datasets of an order of magnitude smaller than our full 18K subject dataset.
Fig.~\ref{fig:genhmc-qual-res} demonstrates the qualitative results generated from GenHMC trained with 18K full data.

\subsection{Universal Facial Encoders with GenHMC Data}
\label{subsec:experiment-ue}
A key downstream application of GenHMC is to provide synchronized dome-HMC correspondences for training universal encoder (UE). As~\citet{bai2024ue} noted, a major challenge is generalization, \textit{i.e.}, models trained on limited paired data tend to overfit to the small set of training subjects.

In contrast, with GenHMC, the generative model offers natural augmentations on various axes: the GenHMC framework enables us to keep the expressions intact while varying orthogonal factors like illuminations, mustache, thickness of the eyebrow, iris color, glasses, etc. in a \emph{controlled} manner. Indeed, we demonstrate that leveraging the GenHMC captures (in addition to some real captures) significantly improve the performance of the current best encoder on held-out validation subjects on metrics across the board, offering substantially more accurate, generalizable and robust facial encoding on unseen test subjects compared to baseline.
Furthermore, it leads to improved scalability with respect to the number of captures.
We highlight these advantages through the following experiments. In Sec.~\ref{subsubsec:ue-comparison}, we present quantitative and qualitative studies against the baseline approach. In Sec.~\ref{subsubsec:genhmc-ue-scaling} and Sec.~\ref{subsubsec:ue-scaling}, we revisit how UE training scales when synthetic data is incorporated.

\begin{table}[tbp!]
\ra{1.3}
\caption{\textbf{Quantitative comparison between UEs.} All models are trained for 200K iterations with a batch size of 16. Note that the encoder trained solely on GenHMC data performs poorly because it computes metrics against the pseudo correspondences from~\citet{wei2019VR}, which biases towards real HMC captures.
See Sec.~\ref{subsubsec:ue-comparison} for more details.}
\resizebox{0.45\textwidth}{!}{
\begin{tabular}{l|cccc}
\toprule
& Photo. $L_1$ $\downarrow$ & Geo. $L_2$ $\downarrow$ & Geo. $L_2$ (Mouth) $\downarrow$ & Lip Shape $\downarrow$ \\
\midrule
Real+GenHMC (Ours) & \cellcolor{green!25}\textbf{2.8598} & \cellcolor{green!25}\textbf{1.2325} & \cellcolor{green!25}\textbf{1.8959} & \cellcolor{green!25}\textbf{1.9968} \\
\midrule
Real (Baseline) & 2.9149 & 1.2603 & 1.9666 & 2.0921 \\
\midrule
Real w/o SSL & 2.9578 & 1.2740 & 1.9811 & 2.1519 \\
\midrule
GenHMC Only & 3.5501 & 1.5118 & 2.4387 & 2.5563 \\
\bottomrule
\end{tabular}}
\label{tab:ue-val-results-log}
\end{table}

\begin{figure}[tbp!]
  \includegraphics[width=0.45\textwidth]{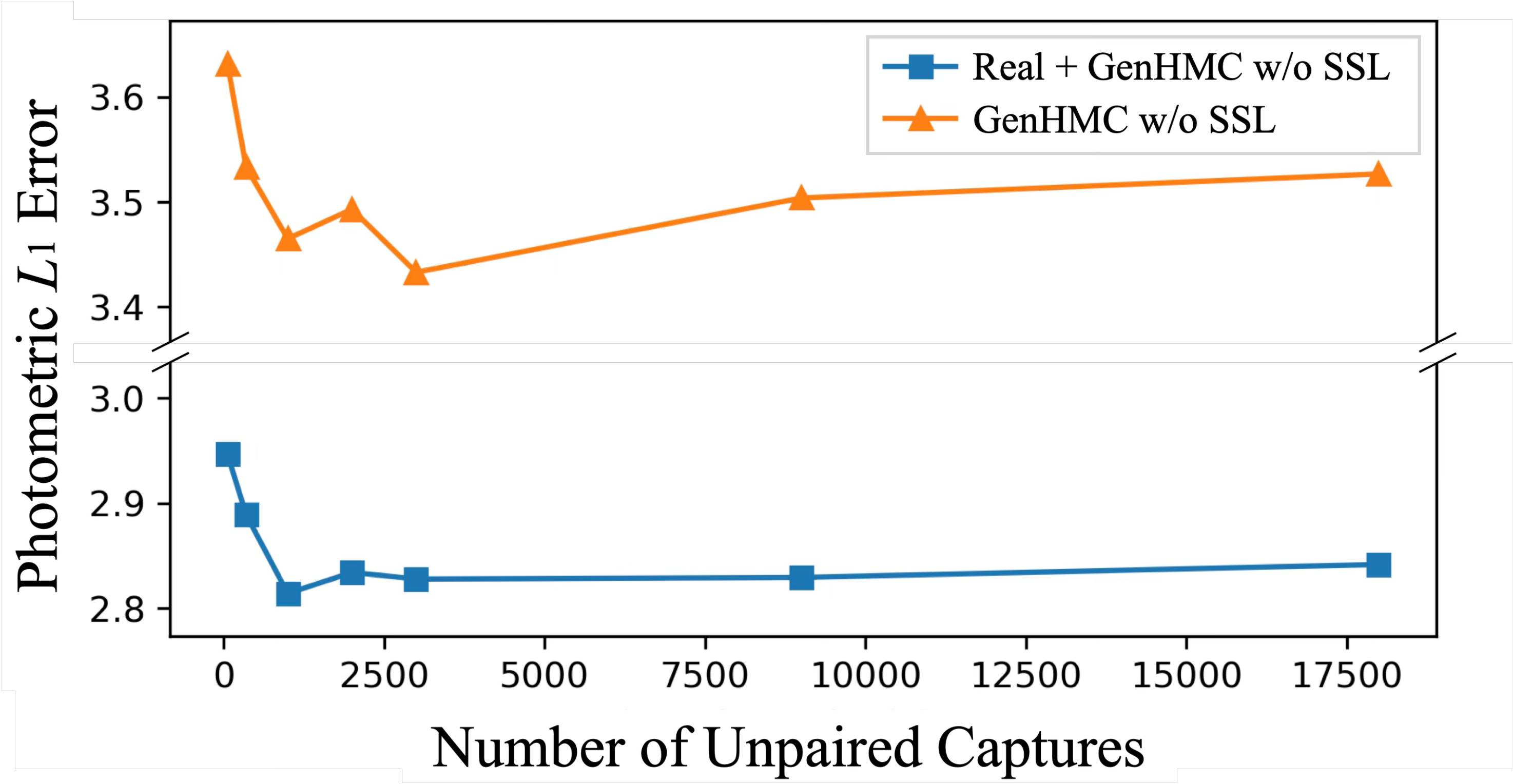}
  \caption{\textbf{Scalability of GenHMC subjects on UE training.}
  Generally, the more the GenHMC training subjects there are, the more accurate the pixel-level reconstruction quality becomes during downstream UE training.
  We use all 307 real subjects together for training and evaluate on held-out 34 subjects.
  To assess the impact of GenHMC data size alone without other priors, we compare encoders trained from scratch without SSL pre-training.
  }
  \label{fig:ue-losses}
\end{figure}

\subsubsection{Comparisons with the Traditional Approach}
\label{subsubsec:ue-comparison}

We conducted quantitative and qualitative studies to compare facial encoders trained with real data and/or GenHMC synthetic data under 3 configurations: (1) mixing (roughly) 50\% real and 50\% synthetic HMC frames, (2) synthetic frames only, and (3) real frames only.

As shown in the quantitative results of Tab.~\ref{tab:ue-val-results-log}, we show that the addition of GenHMC synthetic captures improves the robustness and generalization of the facial encoder while evaluated on held-out unseen test subjects' data across all metrics. Notably, we improve over the current best model by over 5\% on key metrics like the mouth geometry error and the lip shape error. We also specifically note that the encoder trained with synthetic captures only has a significantly higher error, which should be taken with a grain of salt. These errors are computed against the ``ground truths'' which actually are the pseudo correspondences that the method of~\citet{wei2019VR} yields for the real HMC captures. As explained in Sec.~\ref{subsubsec:evaluation-metrics}, this process by itself is prone to bias and errors, which becomes evident when we also train on some real HMC data along with these pseudo labels. In contrast, this encoder is trained with only synthetic HMC images and the actual dome-frame expression latent codes (\textit{i.e.}, perfect ground truth).

For the qualitative evaluations, we also demonstrate a number of frames in the validation set and the corresponding predicted expressions by the aforementioned settings in Fig.~\ref{fig:ue-qualitative-comparisons} (where again, the ``G.T.'' (ground truth) here is actually the pseudo correspondences). In all cases, the mixed setting (ours) produces the most stable and accurate results. The synthetic-only setting also produces expressive and robust encoding, demonstrating an effective learning of the HMC-avatar correspondences captured by GenHMC.

Although we cannot provide proprietary details, training and inference with GenHMC are more than an order of magnitude more efficient than collecting large-scale paired real captures, underscoring its practicality for scalable deployment.
Furthermore, while we adopted a 50-50 sampling weight for real \& synthetic data mix, we note that it is highly likely that a different relative ratio would yield a better outcome, which we leave for future work.

\begin{figure}[tbp!]
  \includegraphics[width=0.95\columnwidth]{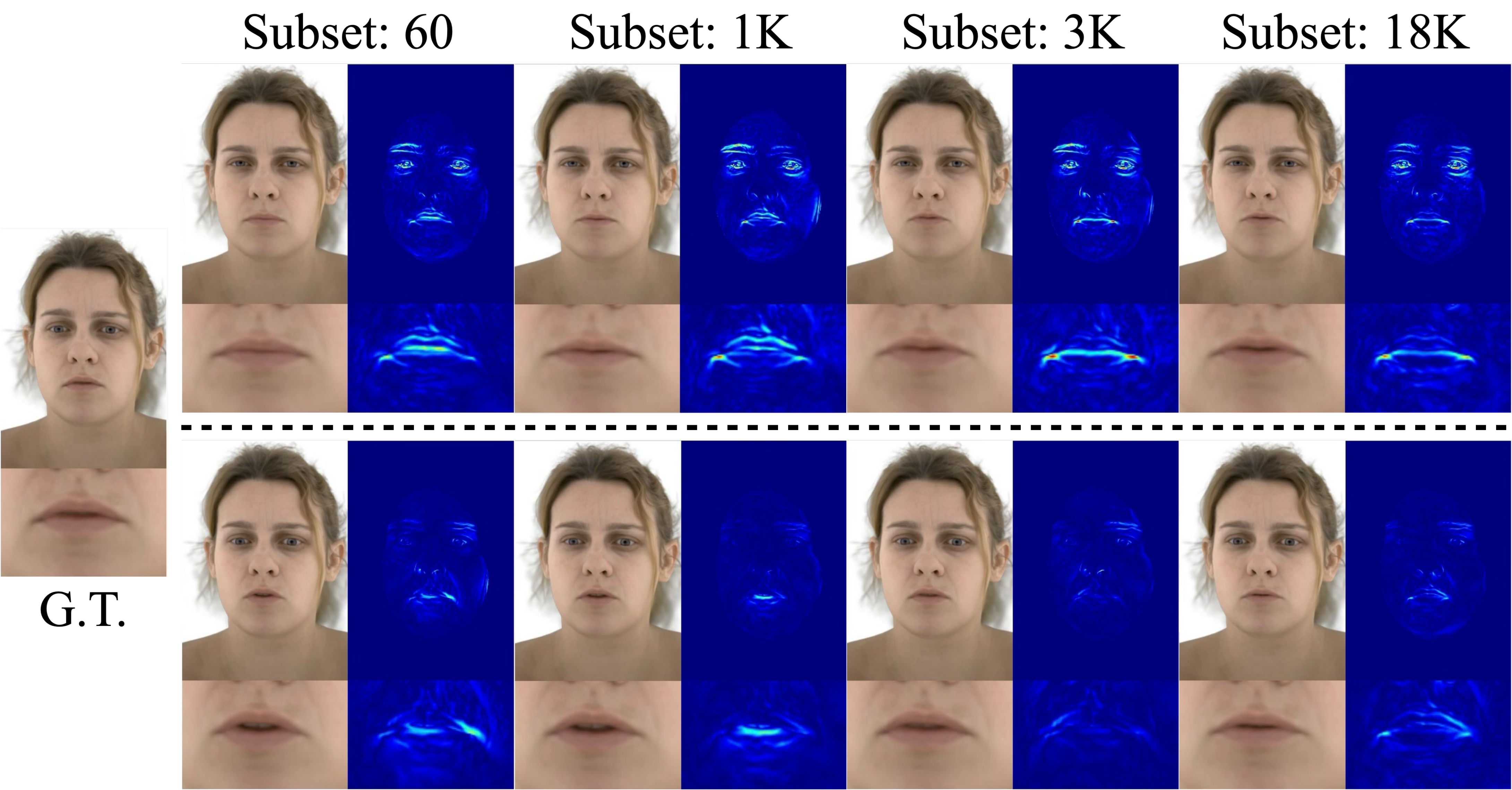}
  \caption{\textbf{Qualitative comparison of GenHMC scalability on UE training.}
  Training UE with only synthetic data (top) improves performance with more diverse GenHMC subjects, while combining real and synthetic data (bottom) yields robust results.
  For each case, the UE driving result is on the left, and the $L_1$ photometric error map (vs. G.T.) is on the right.
  Colors closer to blue indicate lower error.
  Kindly refer to our video material for more results.
  }
  \label{fig:ue-genhmc-scaling-qual-res}
\end{figure}

\begin{figure}[tbp!]
  \includegraphics[width=0.43\textwidth]{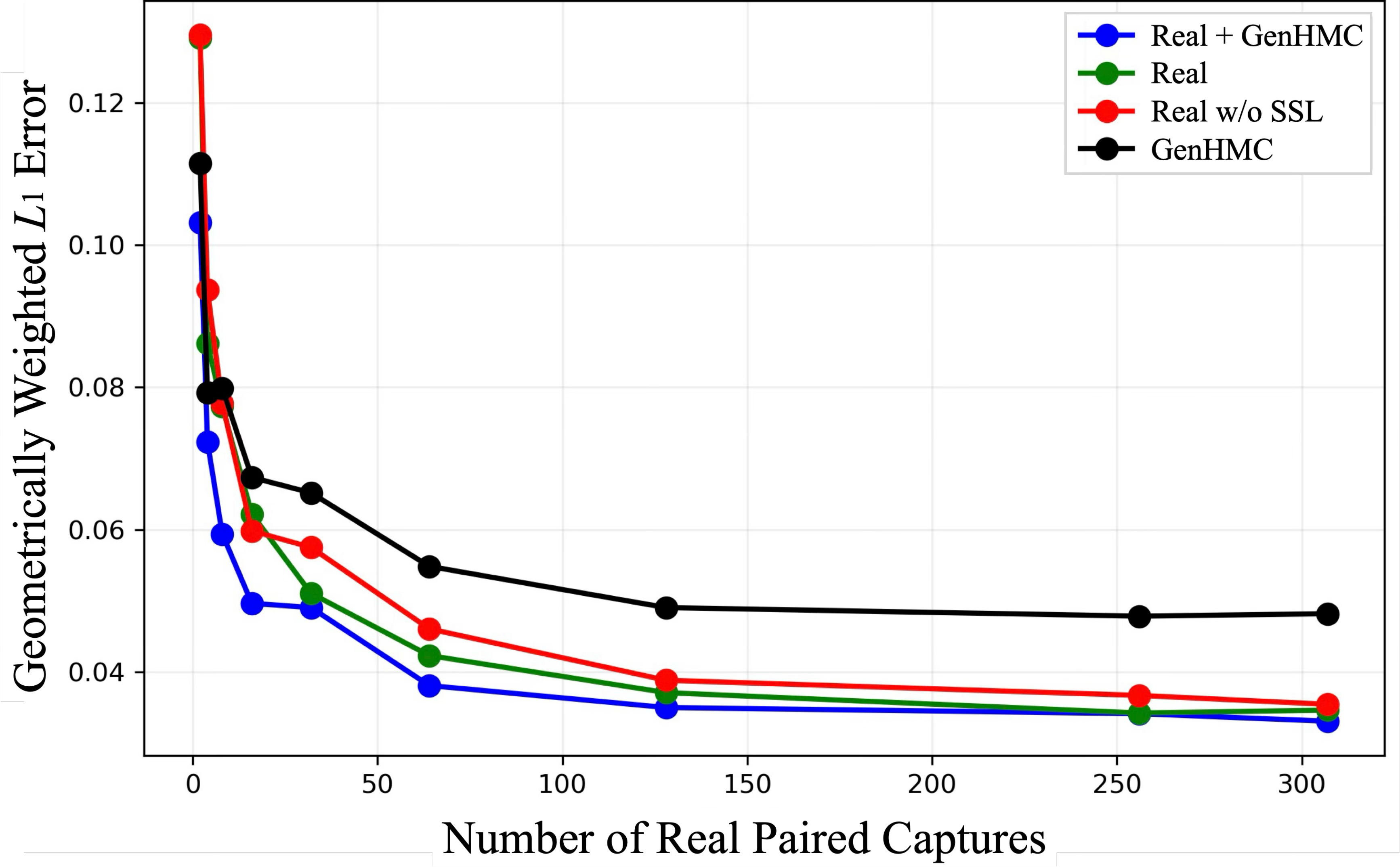}
\caption{\textbf{Scaling law of number of training subjects (real captures) with UE.} Even with a small number of real training subjects, using synthetic data from GenHMC reliably boosts UE performance. We use all 18K subjects for GenHMC data and report the mean of geometrically weighted $L_1$ error, which applies spatial importance weighting, emphasizing critical facial regions through UV mapping and predefined weights.}
  \label{fig:ue-unpaired-losses}
\end{figure}

\subsubsection{Scaling Law of Number of GenHMC Subjects}
\label{subsubsec:genhmc-ue-scaling}
Increasing the number of GenHMC training subjects generally leads to decreased photometric $L_1$ error in the UE training, although this improvement plateaus around 1K to 3K subjects, as shown in Fig.~\ref{fig:ue-losses}.
Note that, to isolate the effect of GenHMC data size on encoder performance without the influence of pre-trained priors, we use encoders that are not initialized with SSL pre-trained weights.
Also as shown in Fig.~\ref{fig:ue-genhmc-scaling-qual-res}, while performance improves with more diverse GenHMC training subjects when training solely on synthetic data (top), combining real and GenHMC images (bottom) yields robust UE performance even with GenHMC models trained on as few as 60 subjects.

\subsubsection{Scaling Law of Number of Subjects with UE}
\label{subsubsec:ue-scaling}
We investigated the impact of the number of paired captures used to train the downstream UE on its performance.
We ablated the number of real captures by training UEs on varying numbers of subjects with maximum of 307 while keeping the number of GenHMC training subjects consistent.
As shown in Fig.~\ref{fig:ue-unpaired-losses}, incorporating synthetic data generated by GenHMC consistently improves UE performance, even with a limited number of real training subjects. This highlights the ability of GenHMC to augment limited real datasets, enhancing the performance and robustness of downstream UE training.

\section{Discussion and Conclusions}
We introduce GenHMC, a generative approach for synthesizing HMC captures to solve the HMC-avatar correspondence problem in training universal face encoders. Unlike prior analysis-by-synthesis methods, GenHMC improves correspondence quality while removing the need for costly paired captures across camera setups. By training diffusion models conditioned on expression-related features, we enable scalable generation of realistic HMC images.
The encoders trained on both real and GenHMC data show improved avatar driving performance and potential for the lower deployment costs, advancing the scalability of photorealistic avatars in VR.

While effective, our method has several limitations. It generates frames independently, lacking temporal consistency, and does not condition on subject identity, limiting identity preservation.
These issues do not impact on our per-frame, identity-agnostic encoder setup, but extending the framework to multi-frame generation and identity-aware conditioning could broaden its applicability.
In addition, the accuracy of the keypoint and segmentation models, as well as the fidelity of avatar renderings, can influence the quality of conditional inputs and thereby the final outputs.
Although our conditioning framework helps mitigate minor errors, these components remain important factors in overall performance.

\begin{acks}
The researchers at SNU were supported by NRF (2021R1A2C3006659) and IITP grants (RS-2021-II211343, RS-2022-II220320), all funded by the Korean Government.
\end{acks}

\newpage

\bibliographystyle{ACM-Reference-Format}
\bibliography{reference}

\begin{figure*}
    \centering
    \includegraphics[width=0.9\linewidth]{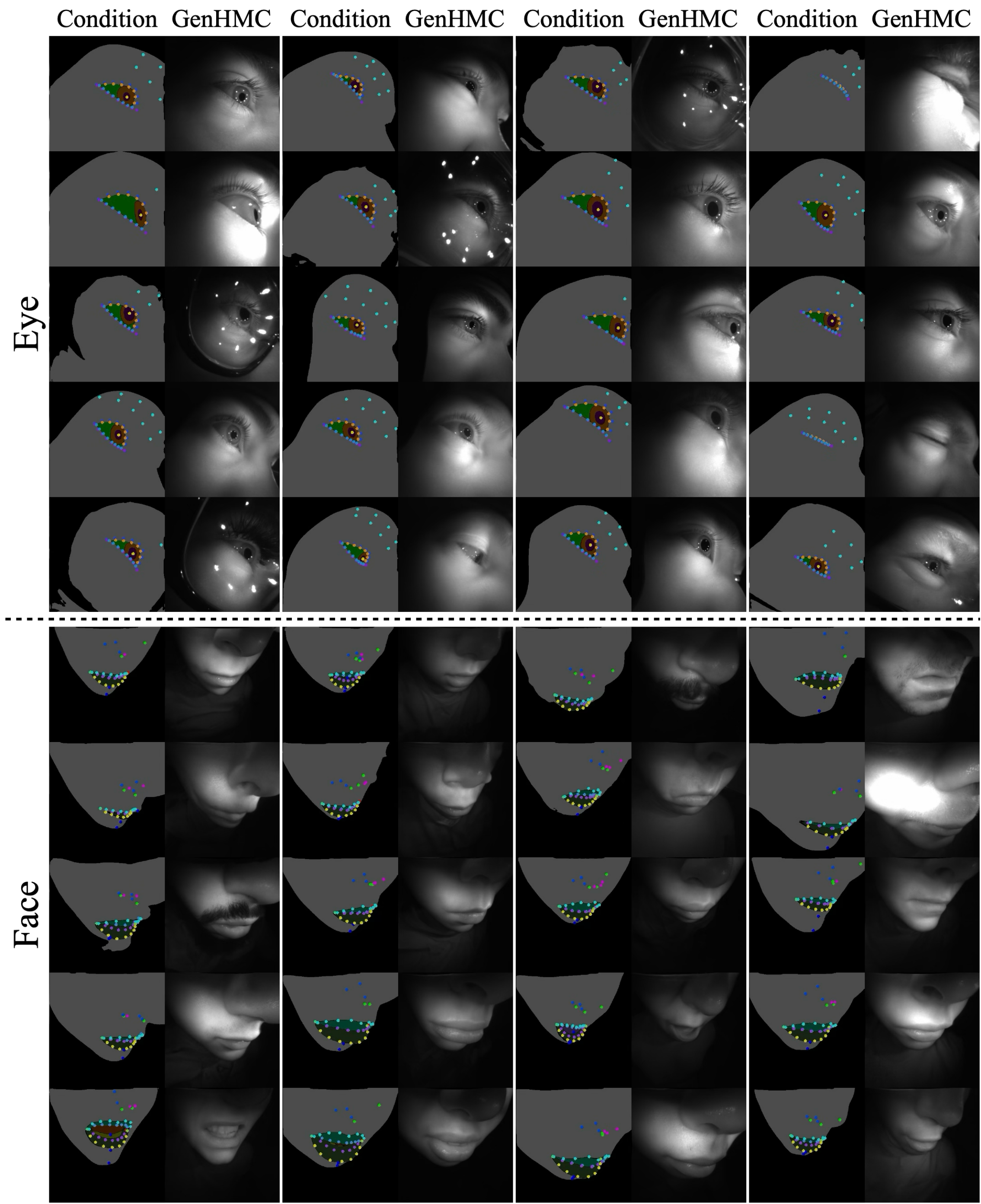}
    \caption{\textbf{Qualitative results of GenHMC.}
    Using a keyseg map as a condition (the left image in each group), our GenHMC generates synthetic HMC images (right) that feature diverse natural augmentations for a single expression.
    }
    \label{fig:genhmc-qual-res}
\end{figure*}

\begin{figure*}
    \centering
    \includegraphics[width=1\linewidth]{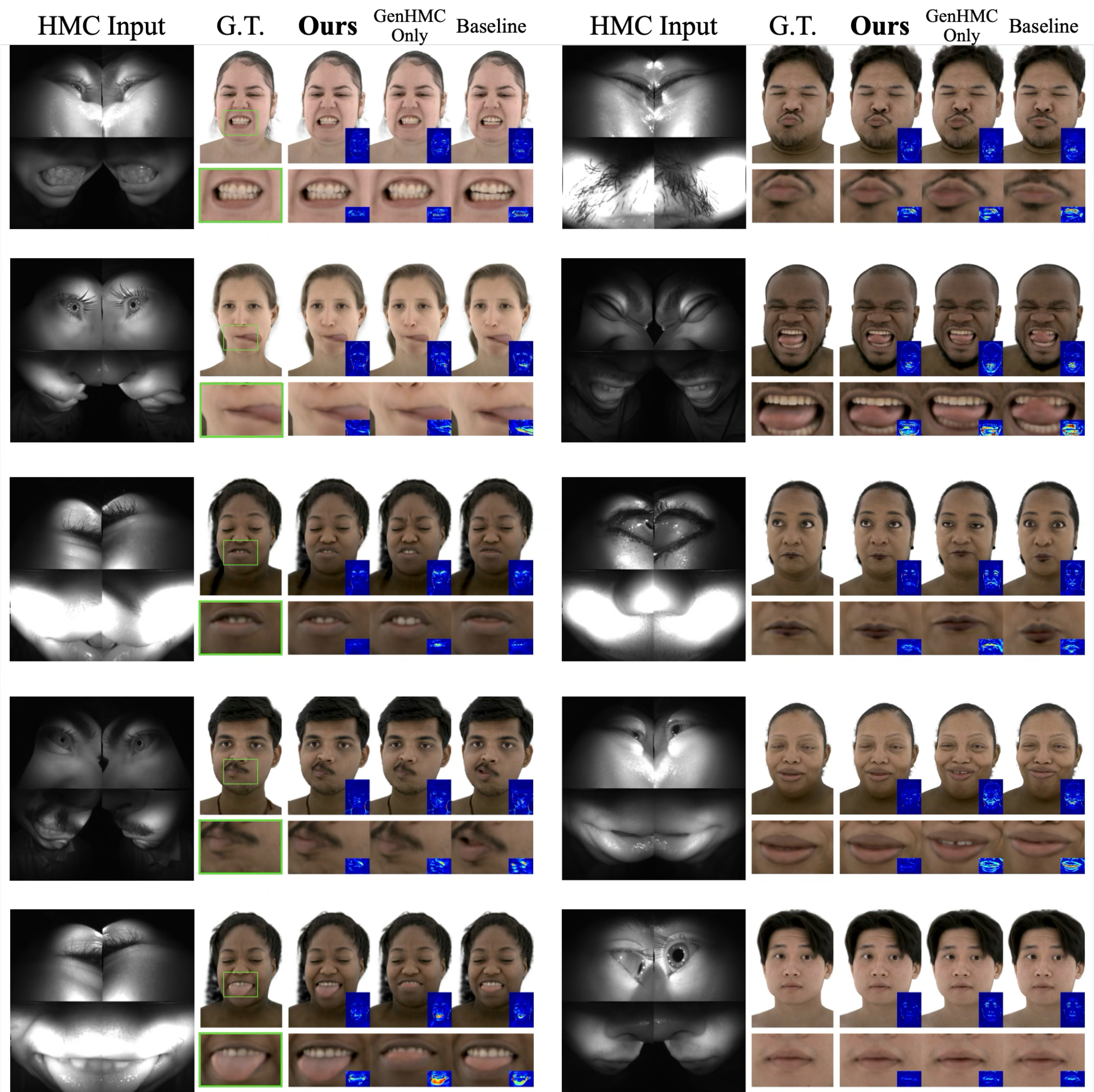}
    \caption{\textbf{Qualitative Comparisons of Universal Encoder (UE).} We present a qualitative comparison of encoders trained with real data and/or GenHMC synthetic data under 3 configurations. For each column, the images are from left to right: HMC input images, ground truth rendering, \textbf{Ours (50\% Real-50\% Synthetic)}, \textbf{GenHMC (Synthetic) Only}, and \textbf{Baseline (Real Only)}~\citep{bai2024ue}. The error map of each encoder's prediction with respect to the ground truth is shown in the bottom-right corner of each rendering. Our results indicate that the mixed real–synthetic training strategy effectively improves inner mouth tracking, capturing subtleties such as tongue movements and teeth closure.}
    \label{fig:ue-qualitative-comparisons}
\end{figure*}

\clearpage
\setcounter{page}{1}
\setcounter{section}{0}
\setcounter{table}{0}
\setcounter{figure}{0}
\renewcommand\thesection{\Alph{section}}
\renewcommand\thetable{\Alph{table}}
\renewcommand\thefigure{\Alph{figure}}

\maketitlesupplementary

We also explore various applications of GenHMC, leading to more controllable framework.
First, Sec.~\ref{supp_sec:glasses_control} explores augmentations involving accessories like glasses, enabling the generation of synthetic HMC captures for a wider range of appearances.
Second, as detailed in Sec.~\ref{supp_sec:multi-view_consistency}, we improve the consistency of synthetic subjects across the different views by leveraging the 3D self-attention mechanism used in multi-view diffusion frameworks and incorporating camera parameters into the diffusion model through timestep embeddings.
Finally, we incorporate light source maps as additional conditional inputs as well as keyseg maps, enabling controllable lighting augmentations (Sec.~\ref{supp_sec:lighting_control}).

\section{Glasses Control}
\label{supp_sec:glasses_control}
\begin{figure}[tbp!]
    \includegraphics[width=1.0\linewidth]{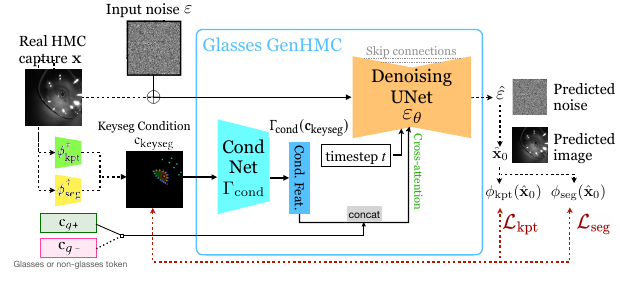}
    \caption{\textbf{Glasses GenHMC training framework.} From Fig. 2, we introduce an additional glasses encoding token $\mathbf{c}_{g\pm}$ corresponding to whether the training sample has glasses or not, concatenated to the keypoint-segmentation condition tokens. At inference time, we can choose to set this token to either $\mathbf{c}_{g+}$ or $\mathbf{c}_{g-}$ to optionally generate glasses in the eye HMCs. Some losses from Fig. 2 are omitted for brevity.}
    \label{fig:glasses_genhmc_overview}
\end{figure}

Wearing glasses under the VR headset device is a common use case, and therefore we want to collect data representing this distribution. However, collecting usable HMC data with subjects wearing glasses is a challenge in itself, and the presence of glasses (and LED reflections on the lenses) also significantly interferes with the person-specific style transfer module that the traditional method of~\citet{wei2019VR} needed to establish HMC-avatar correspondences. Therefore, the traditional approach could not handle glasses cases well, frequently generating inaccurate ``ground truth'' expression codes for the downstream encoder training usage.

In contrast, GenHMC provides an easy solution to this. Specifically, we can extend the regular GenHMC in Sec. 4.1 to be additionally \textit{glasses-conditional}, \textit{i.e.}, \textbf{Glasses GenHMC}, where we generate a synthetic HMC dataset with glasses that comes with accurate, dome-frame ground truth expression codes, provided an initial volume of unpaired HMC captures with and without glasses.

\begin{figure}[tbp!]
    \includegraphics[width=1.0\linewidth]{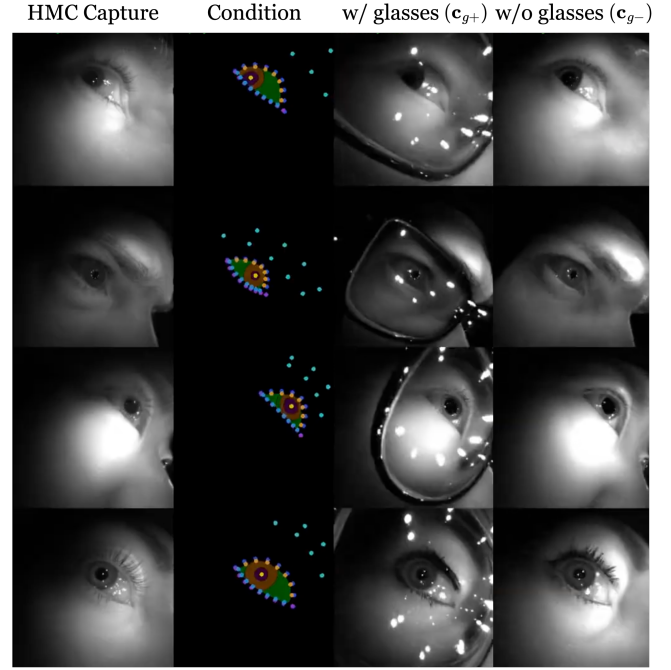}
    \caption{\textbf{Qualitative results of Glasses GenHMC.} Left to right: reference raw HMC image, detected keypoint-segmentation conditional, diffusion with glasses, diffusion without glasses.
    Note that the segmentation map of facial area is not used for Glasses GenHMC to prevent ``cheating'', which could bias the training process towards the semantic leakage of the conditional.} 
    \label{fig:glasses_genhmc_4col}
\end{figure}

To achieve this, we take advantage of a large, unpaired (with dome captures) collection of HMC captures, where frames in the dataset are tagged with binary labels of glasses on or off for each particular capture, $\{g+, g-\}$ (this can be done in an automated way by training a binary classifier as well). The diffusion model is then trained as described before in Sec. 4.1 with the same losses, but the conditional is now a concatenation of the corresponding glasses embedding token $\{\mathbf{c}_{g+}, \mathbf{c}_{g-}\} \in \mathbb{R}^{C}$ (e.g., $C=160$ as in Sec. 4.1) and the flattened feature map $\Gamma_\text{cond}(\mathbf{c})$ for every sample, as illustrated in Fig~\ref{fig:glasses_genhmc_overview}. In this way, the model learns conditioning context from not only the keyseg map informing sparse facial structure, but also a controllable binary token we insert indicating glasses presence. Fig.~\ref{fig:glasses_genhmc_4col} shows that we can successfully generate synthetic HMC data with glasses ($c_{g+}$), while following the same keyseg map as those without glasses ($c_{g-})$.

\begin{figure*}[tbp!]
  \includegraphics[width=0.7\textwidth]{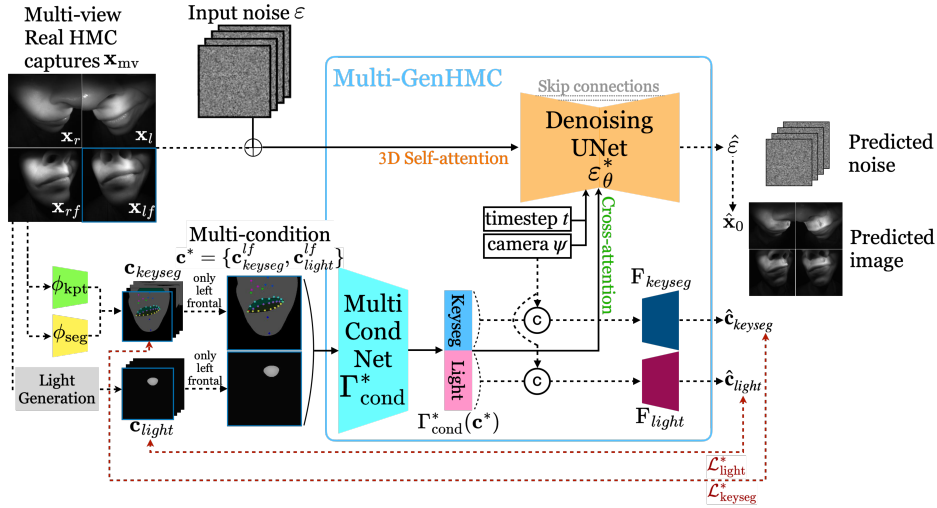}
  \caption{\textbf{Multi-GenHMC training framework}. We leverage the light source map as an additional condition to achieve more precise control over illuminance, and generate multi-view consistent HMC captures by using the camera parameters and 3D self-attention. Note that the training objectives used in regular GenHMC have been omitted for simplicity in the figure. 
  }
  \label{fig:mv_genhmc_overview}
\end{figure*}

\section{Multi-view Consistency}
\label{supp_sec:multi-view_consistency}
For a single subject, HMC captures are collected from multiple different camera views.
However, regular GenHMC uses the detected keyseg map corresponding to a single view as a condition to generate synthetic outputs, which does not guarantee consistency across different views.

\begin{figure}[tbp!]
  \includegraphics[width=0.48\textwidth]{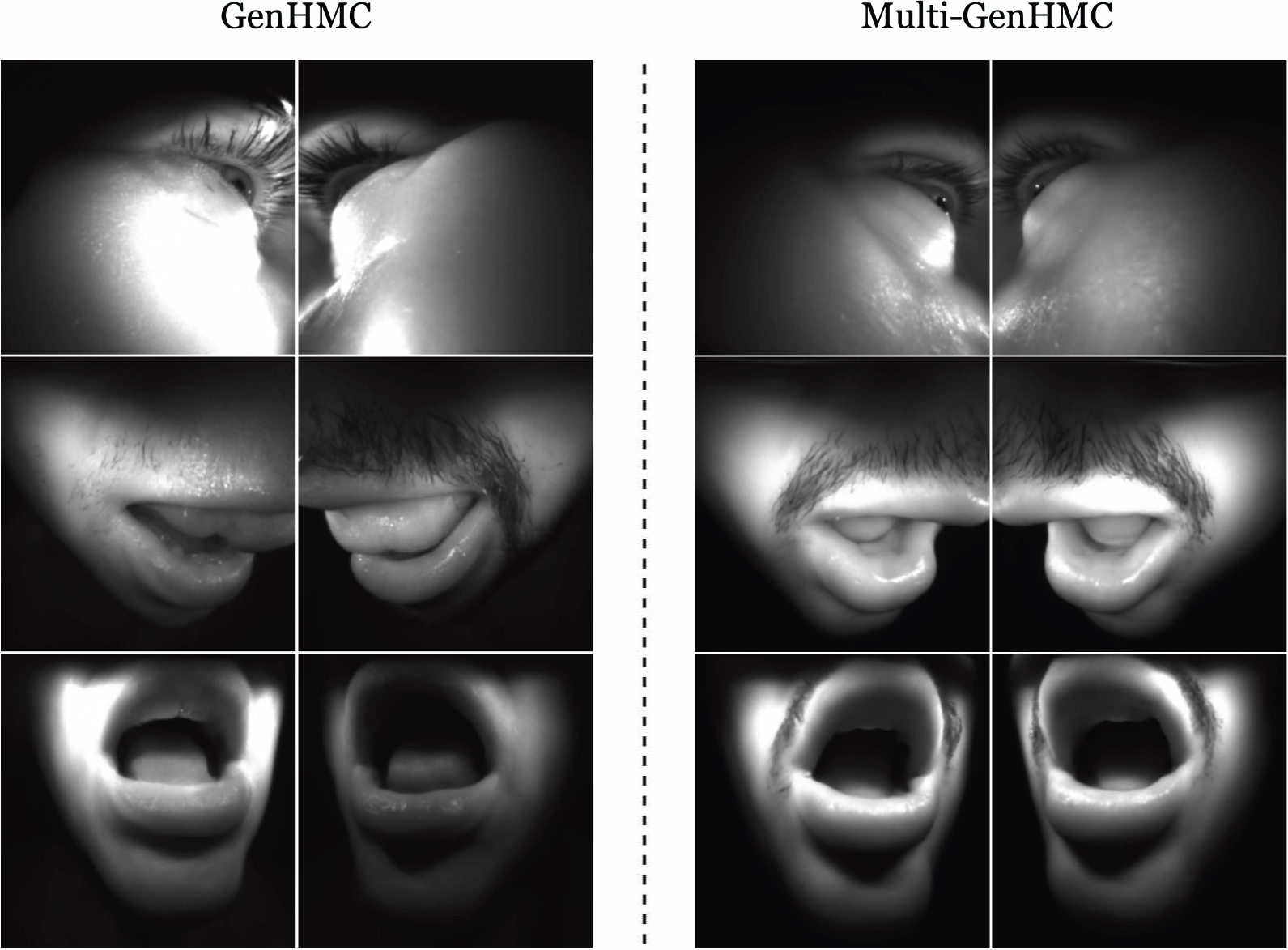}
  \caption{\textbf{Synthetic HMC image examples of GenHMC and Multi-GenHMC}. While GenHMC generates synthetic HMC captures per view and may exhibit different facial details across different views even within the same frame, Multi-GenHMC ensures consistent details for the target views during generation.
  }
  \label{fig:consistency-comparison}
\end{figure}

To address this point, we further conceptualize \textbf{Multi-GenHMC}, which leverages multi-view real HMC data and their camera parameters to generate synthetic captures that consistently share details across different views, as illustrated in Fig.~\ref{fig:mv_genhmc_overview}. 
Inspired by \citep{shi2023mvdream}, a multi-view diffusion framework capable of generating 3D-consistent outputs, we utilize camera parameters and 3D self-attention to produce multi-view consistent synthetic captures.
Given multi-view real HMC captures $\mathbf{x}_\text{mv} = \{\mathbf{x}_{r}, \mathbf{x}_{l}, \mathbf{x}_{rf}, \mathbf{x}_{lf}\}$ consisting of right, left, right-frontal, and left-frontal views, and 12-dimensional camera parameters $\psi$ (comprising 6D rotation~\citep{zhou2019continuity} and camera intrinsic parameters), the camera parameters $\psi$, which are embedded together with timestep $t$, are processed through U-Net $\varepsilon^{*}_\theta$ equipped with 3D self-attention.
As shown in Fig.~\ref{fig:consistency-comparison}, the details of the generated images, e.g., mustache, tongue position, and eyelashes, are much more consistent across different views compared to regular GenHMC.
Note that $\varepsilon^{*}_\theta$ is trained with the same training losses as regular GenHMC.

\begin{figure}[tbp!]
  \includegraphics[width=0.48\textwidth]{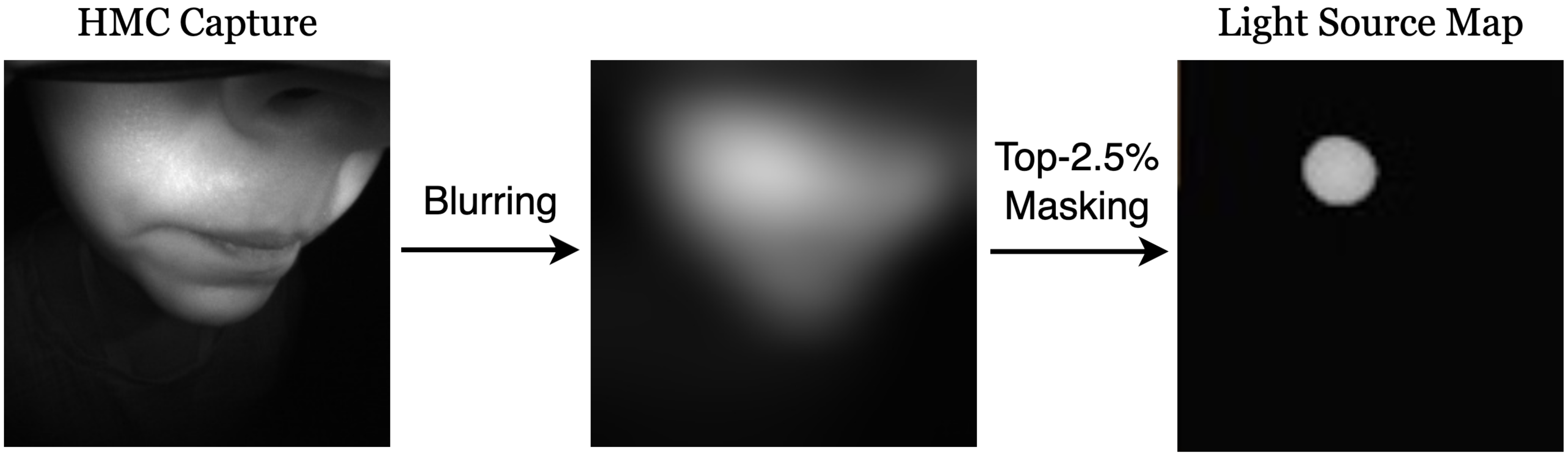}
  \caption{\textbf{Light source map generation}. A light source map, which serves as a conditional signal for lighting, is generated from the input real HMC capture through the blurring and top-value masking process.
  }
  \label{fig:light_source_map}
\end{figure}

\begin{figure*}[tbp!]
  \centering
  \includegraphics[width=0.85\textwidth]{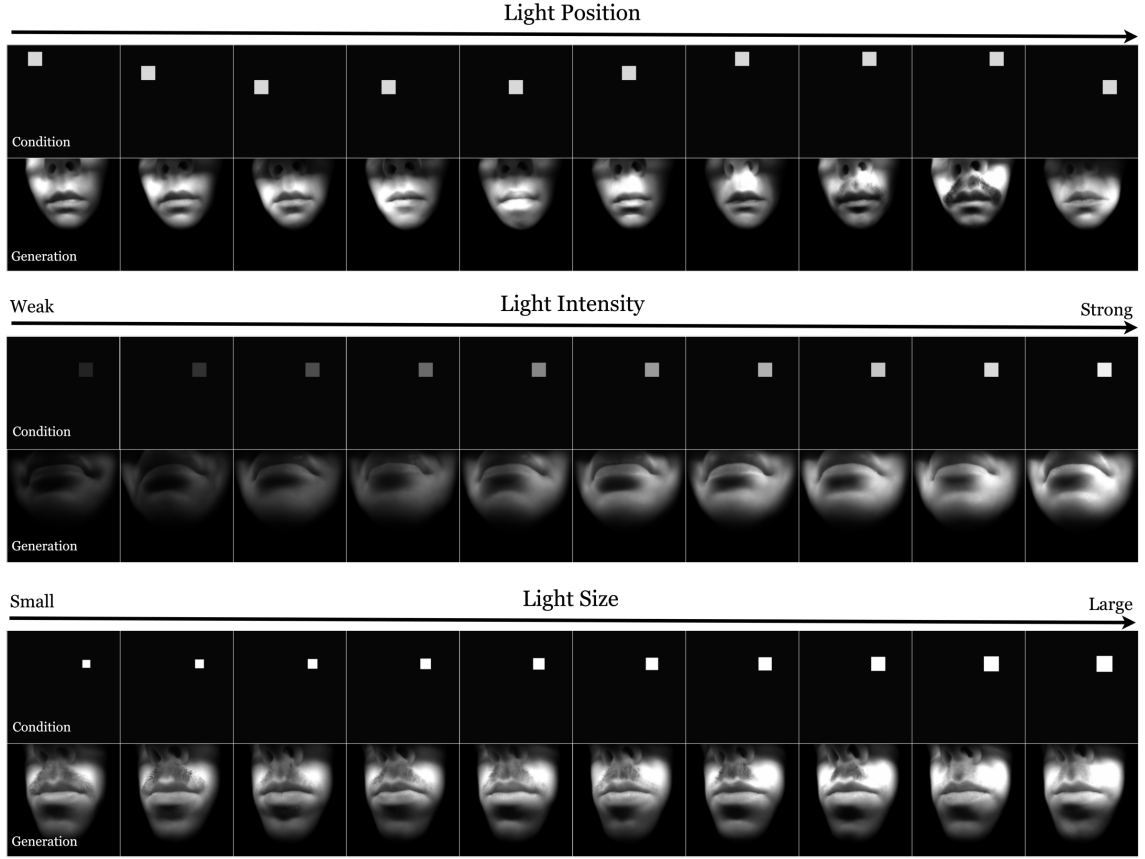}
  \caption{\textbf{Light control with arbitrary light source maps.}
  With our proposed framework, we can generate synthetic HMC images following diverse illuminance conditions, \textit{e.g.}, light position, intensity, and size.
  Note that the identical keyseg maps are used for each row.}
  \label{fig:light_control}
\end{figure*}

\begin{figure}[tbp!]
  \includegraphics[width=0.4\textwidth]{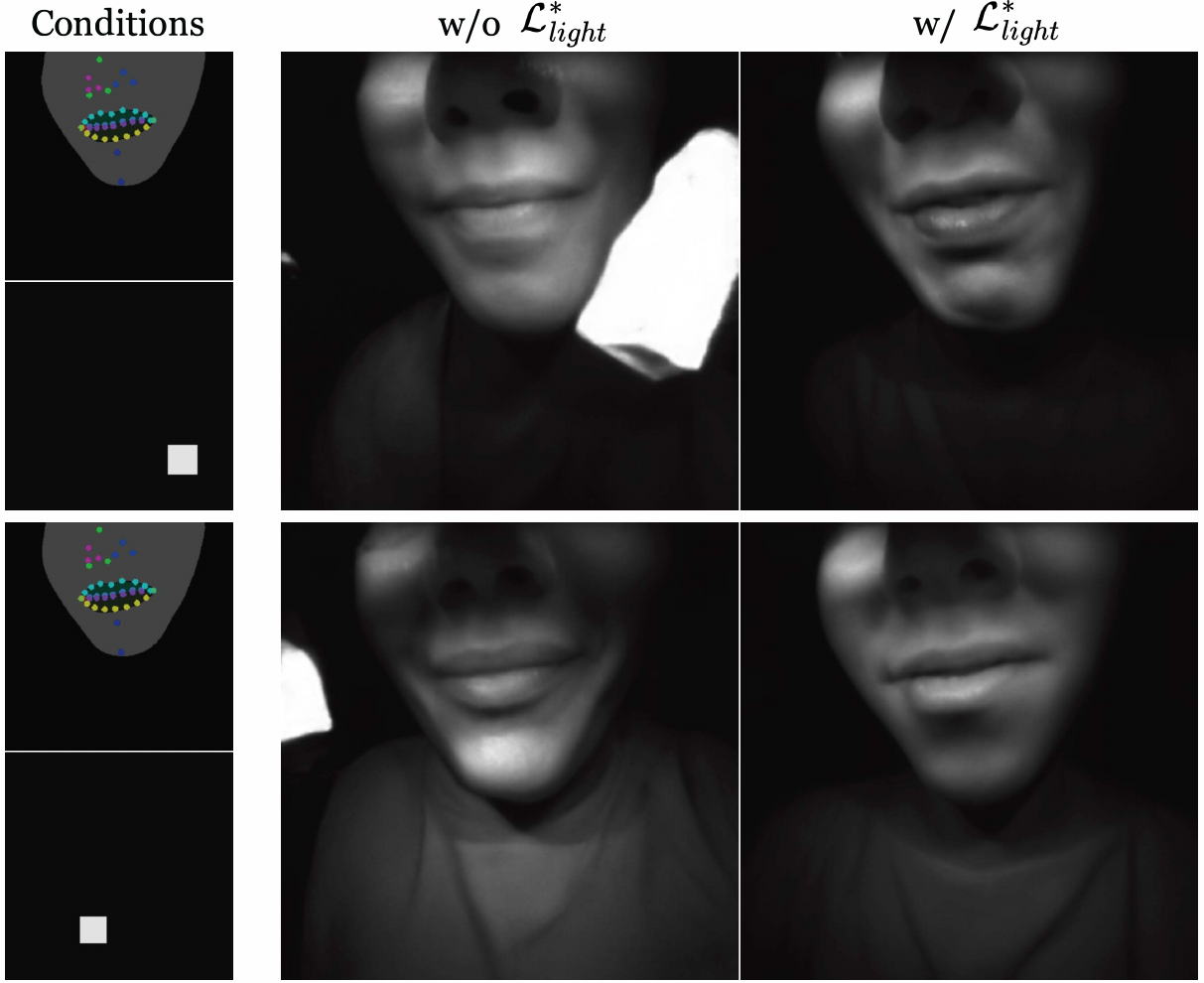}
  \caption{\textbf{Effect of the proposed light reconstruction loss $\mathcal{L}^{*}_{light}$}. Without $\mathcal{L}^{*}_{light}$, more artifacts occur on the background during inference when given the uncommon light conditions, e.g., the light source positioned outside the facial region.
  }
  \label{fig:abl-light-estimator}
\end{figure}

\section{Light Control}
\label{supp_sec:lighting_control}
Another interesting extension of the GenHMC framework is flexible control over the LED illuminations for the synthetic capture. Based on Multi-GenHMC, we incorporate our proposed light source map described in Fig.~\ref{fig:light_source_map} as an additional condition $\mathbf{c}_{light}$, alongside the existing keyseg map $\mathbf{c}_{keyseg}$.

First, the light source map is generated during training by applying a Gaussian kernel to blur the target HMC capture.
This process removes identity-specific details while retaining abstract illumination-related information.
Afterward, we retain only the top 2.5\% of values from the blurred HMC image, corresponding to areas where light intensity is the strongest, to create a point-like representation serving as the final light source map.
By using our proposed light source map as a condition, the model captures the illumination information of the target HMC capture.
The additional light condition allows for finer control over not only the expressions tied to the keyseg map but also the illumination of synthetic HMC captures, providing more flexibility in augmentation.
As illustrated in Fig.~\ref{fig:light_control}, we are able to generate synthetic data with fine controls on light condition.

Furthermore, in a multi-view framework where synthesized views are adjusted based on camera parameters $\psi$, training with all conditions corresponding to each viewpoint can lead to entanglement between the generated image's viewpoint and other conditional inputs.
To prevent this, we set the left-frontal view as the reference view and use only its keyseg map and light source map as input conditions $\mathbf{c}^{*} = \{\mathbf{c}^{lf}_{keyseg}, \mathbf{c}^{lf}_{light}\}$.
Additionally, we introduce additional networks $\mathrm{F}_{keyseg}$ and $\mathrm{F}_{light}$, which take the first and the other half of the conditional feature vector $\Gamma^{*}_\text{cond}(\mathbf{c^{*}}) \in \mathbb{R}^{C \times 64}$
that are processed from $\mathbf{c}_{keyseg}$ and $\mathbf{c}_{light}$, respectively, and camera parameters $\psi$ as inputs and are trained to reconstruct the original conditional inputs for all the target views, including left-frontal:
\begin{equation}
  \mathcal{L}^{*}_{keyseg} = ||\hat{\mathbf{c}}_{keyseg} - \mathbf{c}_{keyseg}||^{2}_{2},
  \label{eq:loss-keyseg-recon}
\end{equation}
\begin{equation}
  \mathcal{L}^{*}_{light} = ||\hat{\mathbf{c}}_{light} - \mathbf{c}_{light}||^{2}_{2},
  \label{eq:loss-light-recon}
\end{equation}
leading to improved quality of the generated outputs dependent on their conditions, e.g., Fig.~\ref{fig:abl-light-estimator} shows the improved output with effectively reduced background artifacts when given unusual $\mathbf{c}_{light}$.

Inference largely follows the same process as regular GenHMC, except that it uses left-frontal avatar rendering to extract keyseg map, camera parameters, and the light source map as additional conditions.
It ensures consistent synthetic outputs while maintaining flexibility in lighting and view conditions.



\end{document}